%% file: main.tex
\documentclass[letterpaper, 10 pt, conference]{ieeeconf}
\IEEEoverridecommandlockouts                                                
\makeatletter
\let\NAT@parse\undefined
\makeatother
\usepackage[colorlinks=true,linkcolor=black,citecolor=blue,urlcolor=blue,breaklinks=true]{hyperref}
\usepackage{graphics}
\usepackage{epsfig}
\usepackage{times}
\usepackage{xcolor}
\usepackage{booktabs,multirow}
\usepackage{amsmath,amssymb,bm}
\usepackage{makecell}
\usepackage{threeparttable}
\usepackage[switch,mathlines]{lineno}
\usepackage{cuted}
\usepackage{capt-of}
\usepackage{colortbl}
\definecolor{lightredbg}{HTML}{FFF5F5} 

\title{\LARGE \bfseries
Adaptor: Advancing Assistive Teleoperation with Few-Shot Learning and Cross-Operator Generalization}

\author{
Yu Liu$^{1}$,
Yihang Yin$^{2}$,
Tianlv Huang$^{1}$,
Fei Yan$^{1}$,
Yuan Xu$^{1}$,
Weinan Hong$^{1}$,\\
Wei Han$^{1}$,
Yue Cao$^{2}$,
Xiangyu Chen$^{2}$,
Zipei Fan$^{1,\dagger}$,
and Xuan Song$^{1}$
\thanks{$^{\dagger}$Corresponding author ({\tt\small fanzipei@jlu.edu.cn})}
\thanks{$^{1}$School of Artificial Intelligence, Jilin University. $^{2}$IO-AI TECH.}
\thanks{This work was partially  supported by the grants of Jilin Provincial International Cooperation Key Laboratory for Super Smart City and Jilin Provincial Key Laboratory of Intelligent Policing.}
}

\begin{document}
\maketitle
\thispagestyle{empty}
\pagestyle{empty}

\begin{strip}\centering
\vspace{-2cm}
\includegraphics[width=\textwidth]{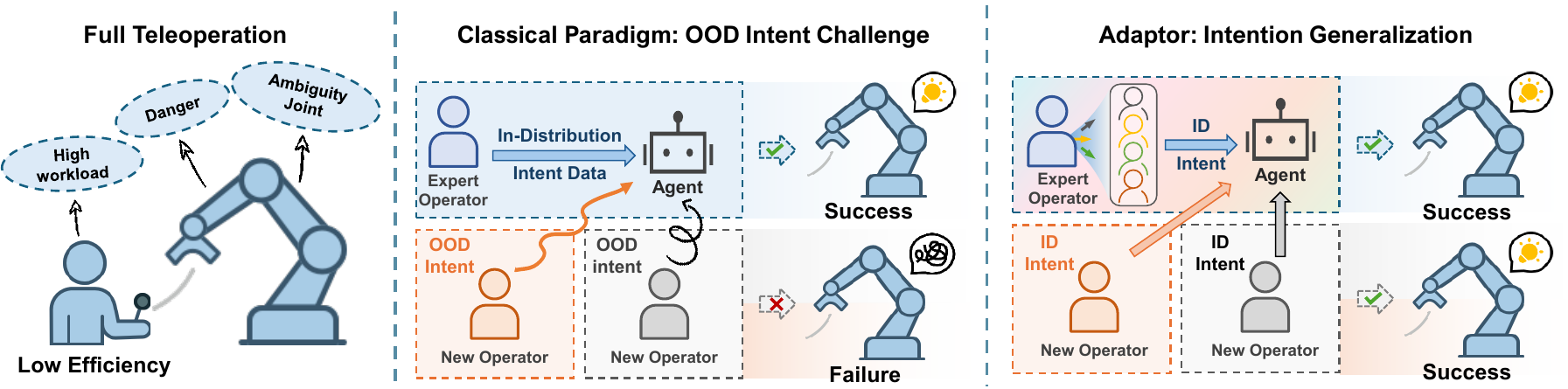}
\captionof{figure}{\textbf{Evolution of teleoperation paradigms.} Left: Direct teleoperation maps human inputs to robot commands but suffers from instability due to human-robot dynamic mismatches. Middle: Conventional assistance relies on expert demonstrations or fixed intent sets, often failing to generalize to diverse operator habits (inter-operator heterogeneity). Right: Adaptor (Ours) models intent uncertainty via trajectory perturbation and geometric keyframes, conditioning a flow-matching VLA policy to robustly adapt to diverse operator behaviors.
\label{fig:feature-graphic}}
\end{strip}

\input{Section/section0_abstract}
\input{Section/section1_introduction}

\input{Section/section2_related_work}
\input{Section/Section3_Problem}

\input{Section/section4_method}

\input{Section/section5_experiment}
\input{Section/section6_conclusion}

\bibliographystyle{IEEEtran}
\bibliography{ref}

\end{document}

%% file: Section/section0_abstract.tex
\begin{abstract}
Assistive teleoperation enhances efficiency via shared control, yet inter-operator variability, stemming from diverse habits and expertise, induces highly heterogeneous trajectory distributions that undermine intent recognition stability. We present Adaptor, a few-shot framework for robust cross-operator intent recognition. The Adaptor bridges the domain gap through two stages: (i) preprocessing, which models intent uncertainty by synthesizing trajectory perturbations via noise injection and performs geometry-aware keyframe extraction; and (ii) policy learning, which encodes the processed trajectories with an Intention Expert and fuses them with the pre-trained vision-language model context to condition an Action Expert for action generation. Experiments on real-world and simulated benchmarks demonstrate that Adaptor achieves state-of-the-art performance, improving success rates and efficiency over baselines. Moreover, the method exhibits low variance across operators with varying expertise, demonstrating robust cross-operator generalization.
\textit{The Homepage is available at:} \textit{\href{https://rainyrobo.github.io/Adaptor/}{\textcolor{magenta}{https://rainyrobo.github.io/Adaptor/}}}.   
\end{abstract}

%% file: Section/section1_introduction.tex
\section{Introduction}

    Teleoperation systems are foundational to advancing embodied intelligence. By leveraging human multimodal perception and expert decision priors, they enable the collection of demonstration data for training and optimizing robotic generalist policies~\cite{black2024pi_0, liu2024rdt,kim2024openvla,zitkovich2023rt}. In typical systems, inverse kinematics (IK) maps operator-specified end-effector poses to joint-space commands in real time~\cite{iyer2024open,Zakka_Mink_Python_inverse_2025}. However, inherent mismatches between human and robot dynamics, particularly under suboptimal operation, can drive joints toward their limits or into singular configurations~\cite{10.1109/TRO.2024.3431830}. Consequently, reducing operator workload while improving teleoperation efficiency and quality remains a central research objective~\cite{luo2024human, yoneda2023noise,liu2024dragon}.

    Assisted teleoperation typically adopts a shared human-robot control paradigm: the user specifies high-level intent, while an autonomous robot takes over low-level control to improve the efficiency and quality of teleoperation~\cite{padmanabha2024independence, brooks2019, selvaggio2021autonomy}.  Existing work on user intent inference and execution falls into two categories: (i) methods utilizing predefined intent sets and policy libraries to match user goals with specific behaviors~\cite{brooks2019,10.1145/3359614,javdani2018shared,jonnavittula2021know,newman2022harmonic}, and (ii) data-driven models that map high-dimensional user inputs to low-dimensional, task-specific controls~\cite{yoneda2023noise,chen2022asha,zhao2024conformalized,cui2023no,Jonnavittula2022LearningTS,zurek2021situational,jonnavittula2024sari}. Despite their effectiveness, both classes of methods rely on trajectory modeling and therefore struggle to generalize intent recognition across operators and skill levels. Even under identical task conditions, differences in user experience and operational routines induce substantial heterogeneity in the distribution of teleoperation trajectories. As a result, existing methods either expand intent-policy libraries or retrain models for each operator, causing data requirements to scale combinatorially with ``$\text{intent} \times \text{operator style} \times \text{scenario}$'' and thereby significantly increasing the costs of data collection and annotation.

    To address these limitations, we propose Adaptor, an assistive teleoperation framework designed to facilitate few-shot adaptation and robust intent recognition across operators with varying experience levels and behavioral profiles. The system operates through a three-stage pipeline. First, Adaptor injects stochastic noise into demonstration trajectories to form a perturbation distribution, thereby simulating a diverse spectrum of operator behaviors. Second, it employs a keyframe-based intent extraction mechanism to distill critical temporal information, enhancing both the robustness and computational efficiency of intent recognition. Finally, the system encodes these processed trajectories into latent embeddings that condition a Vision-Language-Action (VLA) controller. This controller is trained to recover stable intent representations from perturbations, enabling precise, intent-conditioned policy execution. Our main contributions are summarized as follows:
        \begin{itemize}
            \item We propose Adaptor, a VLA-based shared control framework that balances human intent robustness with VLA efficiency to achieve few-shot cross-operator generalization and reliable task execution.
            \item We introduce an intent modeling pipeline to tackle cross-operator distribution shift. By synthesizing a perturbation distribution and keyframe extraction, our approach alleviates covariate shift and facilitates the few-shot intent modeling of multi-user intents within an end-to-end policy.
            \item Across six tasks in simulation and real-world settings on multiple robotic platforms with participants of diverse operational experience, Adaptor achieves state-of-the-art performance, surpassing pure and assisted teleoperation baselines in success rate, completion time, and user satisfaction.
        \end{itemize}

%% file: Section/section2_related_work.tex
\section{Related Work}

    \subsection{Assisted Teleoperation}
        Assisted teleoperation offers a pragmatic trade-off between human control and robotic autonomy: it preserves the operator's decision-making flexibility while improving task efficiency~\cite{luo2024human,yoneda2023noise,padmanabha2024independence, liu2024dragon}. In this paradigm, the operator performs high-level planning and communicates intent, whereas the robot handles low-level closed-loop control and, conditioned on the parsed intent, executes concrete actions~\cite{chen2022asha,selvaggio2021autonomy,gu2023rt,sundaresan2023rt}. Consequently, the accuracy of intent recognition is a key determinant of overall system performance~\cite{hoffman2024inferring,liu2025casper}. Existing approaches fall into two broad categories. The first comprises retrieval-based methods, which collect operators' intents and their associated low-level actions in advance and, at inference time, retrieve the most similar intent-policy pair to assist teleoperation~\cite{brooks2019,10.1145/3359614,javdani2018shared,jonnavittula2021know,newman2022harmonic}. These methods can further leverage pre-trained Vision-Language models (VLMs)~\cite{team2025gemma} to improve intent inference and align skills more precisely within a library~\cite{liu2025casper}. The second comprises data-driven methods, which jointly learn intent representations and control policies from demonstrations, enabling end-to-end modeling of the intent-policy mapping~\cite{yoneda2023noise,chen2022asha,zhao2024conformalized,cui2023no,Jonnavittula2022LearningTS,zurek2021situational,jonnavittula2024sari}. 

        However, existing methods face significant challenges in cross-operator generalization. Given the long-tailed distribution of operator behaviors and habits, relying solely on expanding skill libraries or accumulating training data fails to cover these highly personalized and sparse features exhaustively. To address this bottleneck that data scaling alone cannot resolve, we propose a personalized adaptation system based on few-shot learning.

    \subsection{Noise Injection to Increase Robustness}
        Learning assistive teleoperation policies is challenging due to the complex, nonlinear mapping between high-level human intentions and robot dynamics. Standard behavior cloning often suffers from covariate shift, where minor deviations from expert demonstrations induce compounding errors and hazardous states~\cite{mehta2025stable,baek2024unexplored}. Drawing on the principle of persistent excitation, researchers can enhance policy robustness and generalization by ensuring that the training data provide broad coverage of the state-action space~\cite{liu2024robust}, or by injecting isotropic Gaussian noise into control inputs to induce diverse perturbations~\cite{GREEN1986351}. A prevalent strategy involves iteratively executing the robot's current policy while requesting supervisor corrections for the visited states; such online human corrections mitigate policy-induced distribution shift~\cite{ross2010efficient,ross2011reduction,wang2025inference,kelly2019hg}. However, this approach imposes a substantial burden on the supervisor and risks exposing physical hardware to suboptimal or hazardous states. Alternatively, noise injection during data collection can yield diverse demonstrations, with the noise magnitude tuned to approximate the trained policy's error profile (e.g., DART)~\cite{laskey2017dart}.
        
        Shifting the focus from pure policy correction to user adaptability, we inject stochastic noise into the supervisor policy to simulate a spectrum of teleoperator behaviors. This strategy effectively improves generalization to previously unseen operator behaviors.

    \subsection{VLA for Robotics}
       Vision-Language models (VLMs) pre-trained on internet-scale data have garnered significant attention for their strong generalization and adaptability across diverse applications~\cite{radford2021learning, li2023blip, gao2023llama}. Building on these advances, Vision-Language-Action (VLA) systems leverage the semantic priors of pre-trained VLMs to facilitate end-to-end robotic control. This integration enables the development of generalist robot policies capable of adapting to diverse robot embodiments and manipulation tasks~\cite{black2024pi_0,kim2024openvla,zitkovich2023rt,liu2025hybridvla, intelligence2025-pi05}.

       While the autonomous execution capabilities of VLA systems offer the potential to alleviate data collection burdens, current models often lack sufficient generalization and execution stability. To bridge this gap, we propose a VLA-based shared control framework for assistive teleoperation, designed to balance the robustness of human strategies with the efficiency of VLA execution.

%% file: Section/Section3_Problem.tex
\section{Problem Formulation} 

    Assistive teleoperation entails both low-level motor control and high-level intent inference. We formalize this process as a Partially Observable Markov Decision Process (POMDP) augmented with a latent intent space, defined by the tuple $\langle \mathcal{S}, \mathcal{A}, \mathcal{O}, \mathcal{P}, \mathcal{Z} \rangle$. Here, $\mathcal{S}$ denotes the unobservable physical state space, $\mathcal{A}$ the action space, and $\mathcal{O}$ the observation space. At each time step $t$, the robot receives a composite observation $o_t \in \mathcal{O}$, defined as the tuple $o_t = (s^{prop}_t, o^{vis}_t, L)$. Specifically, $s^{prop}_t$ represents the fully observable proprioceptive state (e.g., gripper status and joint angles), $o^{vis}_t$ denotes the high-dimensional visual observation (RGB images), and $L$ is the natural-language instruction. $\mathcal{P}: \mathcal{S} \times \mathcal{A} \to \Delta(\mathcal{S})$ defines the transition dynamics. The intent $z \in \mathcal{Z}$ is a latent variable modeling the trajectory-level objective, inferred from the human teleoperation trajectory $\xi^{h}=\{(o_t^{h}, a_t^{h})\}_{t=1}^T$ of horizon $T$. Consequently, the execution of policy $\pi_{\theta}(a_t \mid o_t, z)$ is conditioned on the robot's observations and the inferred operator intent.

    Formally, our goal is to learn a policy $\pi_{\theta}$ that approximates an expert policy $\pi_{\theta^{*}}$. We analyze the performance gap by decomposing the expected loss into two distinct components:    
        \begin{equation}
        \label{eq:loss}
            \begin{aligned}
                \mathcal{L}_{r}(\theta)
                &= \underbrace{
                  \mathbb{E}_{p\left(\xi | \pi_{\theta}, z \right)} \mathcal{J}(\theta,\theta^{*} | \xi)
                 -\mathbb{E}_{p\left(\xi | \pi_{\theta^{*}}, z\right)} \mathcal{J}(\theta,\theta^{*} | \xi)
                }_{\text{distribution shift}}
                \\
                &\quad + \underbrace{\mathbb{E}_{p(\xi| \pi_{\theta^{*}},z)}  \mathcal{J}(\theta,\theta^{*} | \xi)}_{\text{supervised loss}},
            \end{aligned}
        \end{equation}
    where $p(\xi | \pi, z)$ denotes the trajectory distribution induced by executing policy $\pi$ conditioned on intent $z$, and $\mathcal{J}(\theta,\theta^{*}|\xi)$ measures the discrepancy between $\pi_{\theta}$ and $\pi_{\theta^{*}}$ along trajectory $\xi$. The \textit{supervised loss} represents the standard imitation objective on expert demonstrations. Crucially, the \textit{distribution shift} term quantifies the mismatch between the state distributions visited by the learner versus the expert. In teleoperation, this shift is exacerbated by the high variance of human trajectories $\xi^{h}$, stemming from both inter-operator expertise differences and intra-operator stochasticity. Consequently, out-of-distribution (OOD) intents can lead to compounding errors, causing the learned policy to diverge significantly from the expert distribution.
    
    Accordingly, our objective is twofold: (i) to enhance the generalization of intent recognition across varying operator expertise and conditions, thereby mitigating distribution shift; and (ii) to refine policy learning for high-fidelity alignment with expert demonstrations, minimizing the supervised imitation loss.

%% file: Section/section4_method.tex
\section{Method}

    \subsection{Overview}
        As illustrated in Fig.~\ref{fig:framework}, the Adaptor framework operates through two distinct phases: \textit{Preprocessing} and \textit{Policy Learning}. During preprocessing, we construct perturbation distributions over teleoperation trajectories and extract keyframes to approximate intent uncertainty, thereby enhancing recognition robustness. In the subsequent training and inference stages, the pipeline orchestrates three core components: a VLM backbone, an Intention Expert, and an Action Expert. Initially, the VLM processes multimodal inputs to extract environmental context. These representations are then utilized by the Intention Expert, which infers latent intentions by fusing trajectory guidance (sourced from preprocessed trajectories during training or coarse human demonstrations during inference) with the scene's semantic context. Finally, the Action Expert integrates these intent predictions with multimodal embeddings to generate comprehensive and precise robot control commands via flow matching.
        \input{Figure/framework}

    \subsection{Intention Preprocessing}
        \label{sec:Intention_Preprocessing} 
        This subsection addresses the problem of intent \textit{distribution shift} in the error term of Eq.~\ref{eq:loss} for assisted teleoperation. We introduce two mechanisms: an intent perturbation distribution and a keyframe extraction method to model inter-operator variability in intent and to extract salient intents that suppress local perturbations, thereby mitigating errors induced by the \textit{distribution shift}.

    \subsubsection{Intent Perturbation Distribution}
        Treating the supervised teleoperation trajectory $\xi^*$ as a direct proxy for the operator's intent $\xi$ integrates seamlessly into standard policy-learning pipelines and obviates additional data collection. However, behavior cloning under this degenerate intent prior exposes the learner only to the expert-induced state distribution, leading to covariate shift at deployment and limited generalization to unseen variations in operator intent.

        Inspired by the idea of enhancing policy robustness by injecting noise into expert demonstrations~\cite{laskey2017dart}, we perturb supervised teleoperation trajectories to construct an intent perturbation distribution that captures inter-operator variability. Concretely, let a supervised demonstration be the teleoperation trajectory $\xi^*=(a_1^{*},a_2^{*},\dots,a_T^{*})$ collected under an expert policy $\pi_{\theta^{*}}$. We model intent variability with a trajectory-level perturbation kernel $q_\psi(\tilde{\xi}\mid \xi^*)$ and adopt a narrow Gaussian tube around $\xi^*$:
        \begin{equation}
            \tilde{\xi} \sim q_{\psi}(\tilde{\xi}\mid \xi^*) \;=\; \mathcal{N}\!\left(\tilde{\xi};\, \xi^*,\, K_{\psi}\right),
        \end{equation}
        where $K_\psi=\mathrm{diag}(\Lambda_1,\ldots,\Lambda_T)$ is taken in block-diagonal
        form, corresponding to independent per-time perturbations. 
 
        From this distribution we draw intent instances with varying perturbation magnitudes and apply behavior cloning via supervised action regression to approximate the expert policy, thereby implicitly recovering the underlying intent from the perturbed trajectories. This procedure expands the state-action coverage during training without relying on a large corpus of suboptimal or failed examples, improving robustness to intent drift and out-of-distribution (OOD) scenarios.

    \subsubsection{Intent Keyframe Extraction}
        Trajectory segments that are well-approximated by linear interpolation often contain redundant kinematic information regarding the operator's intent. As illustrated in the keyframe extraction module of Fig.~\ref{fig:framework} for a ``pen organization'' task, intent is semantically concentrated at subtask boundaries—such as reaching (initiation), grasping (interaction), and releasing (completion)—rather than in the smooth intermediate transition phases. Consequently, preserving every timestep introduces unnecessary computational overhead. Conversely, naive reduction strategies, such as uniform sampling or pooling, lack the geometric sensitivity required to preserve high-frequency control details, potentially missing critical action keyframes such as the exact moment of interaction.

\begin{figure}[t]
                \centering
                \includegraphics[width=0.85\linewidth]{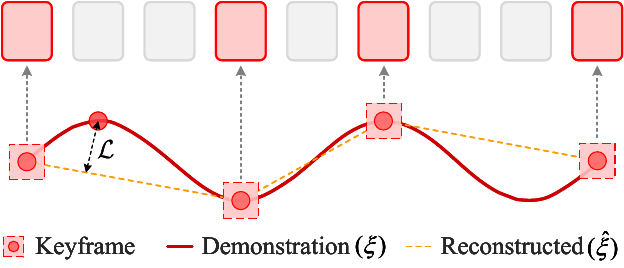}
                \caption{Schematic of the intent keyframe extraction.}
                \label{fig:Intent_Keyframe_Extraction}
            \end{figure}

        Inspired by~\cite{10035484}, we formulate intent extraction as a geometry-aware trajectory compression problem. As illustrated in Fig.~\ref{fig:Intent_Keyframe_Extraction}, given a demonstration trajectory $\xi=\{x_t\}_{t=0}^{T}$, we measure the approximation error of a candidate keyframe subsequence $W$ using the Directed Hausdorff distance. Specifically, for the linear interpolation $\hat{\xi}=f(W)$, the error is defined as:
        \begin{equation}
            \mathcal{L}(\hat{\xi},\xi) \;=\; \max_{x\in \xi} \; \min_{\hat{x}\in \hat{\xi}} \; \ell(x, \hat{x}),
        \end{equation}
        where $\min_{\hat{x} \in \hat{\xi}} \ell(x, \hat{x})$ represents the geometric distance from a raw point $x$ to the continuous piecewise-linear trajectory $\hat{\xi}$. We solve for the minimal cardinality set $W$ (satisfying $0=t_0<\cdots<t_L=T$) subject to the constraint $\mathcal{L}(\hat{\xi},\xi) \le \eta$, thereby ensuring that the worst-case geometric deviation remains within the prescribed tolerance $\eta$.

    \subsection{VLA-Based Assistive Policy Learning}
    \label{sec:assistive_policy_learning}

        \subsubsection{Architecture Implementation} 
        
        Adaptor adopts a Mixture-of-Transformer architecture integrating three components: a VLM and an Action Expert, both inheriting initial weights from the pre-trained VLA model~\cite{black2024pi_0}, and a novel decoder-only Intention Expert, implemented in the same manner as the Gemma backbone~\cite{team2025gemma}.
        
        The VLM is designed to robustly align the natural language instruction $L$ with visual observations, denoted as $o^{vis}_t = \{I_t^{i}\}_{i=1}^{N}$. At each timestep $t$, a SigLIP encoder extracts high-level perceptual features from $o^{vis}_t$, which are subsequently integrated with language tokens within a shared Transformer space to facilitate deep cross-modal fusion.  This architecture explicitly captures the semantic correspondence between task goals and the scene context, yielding consistent representations that serve as a reliable foundation for downstream intent understanding and action execution.
        
        The Intention Expert is designed to model the operator's latent intent $z$. To enhance robustness against incomplete or interrupted demonstrations encountered during inference, we employ a temporal truncation strategy during training, where expert trajectories are randomly cropped at the end. The preprocessed teleoperation trajectories are first projected into the language embedding space via a linear layer. These trajectory tokens are then concatenated with the VLM features to form a unified context, enabling bidirectional attention for deep semantic alignment and intention recognition.

        The Action Expert translates multimodal context into action chunks $A_t = (a_t, a_{t+1}, \dots, a_{t+H})$. We map the proprioceptive state $s^{prop}_t$ and a flow-matching noise term $\epsilon$ into the action embedding space to form queries. These queries drive a cross-modal attention decoder, which attends to the multimodal context (language $L$, visual observations $o^{vis}_t$, and intention $z$) serving as Keys and Values. This architecture ensures that the generated actions are strictly conditioned on both the semantic environment and the specified intention constraints.
    
        \subsubsection{Training Objective}
            We adopt a hybrid training strategy to instantiate the policy. To leverage the foundational priors of the pre-trained VLM and Action Expert, we apply Low-Rank Adaptation (LoRA)~\cite{hu2022lora}. Conversely, the Intention Expert undergoes full-parameter training to ensure alignment with the VLM context. Building upon these integrated representations, the policy $\pi_\theta$ employs Conditional Flow Matching (CFM) to model the continuous transformation from a Gaussian prior to the expert action distribution. Specifically, we construct the interpolated action $a_t^\tau = (1-\tau)\epsilon + \tau a_t$ and train $\pi_\theta$ to regress the target vector field $v(a_t^\tau, \tau) = a_t - \epsilon$ via the following objective:
            \begin{equation}
                \mathcal{L}_{\text{action}} = \mathbb{E}_{t, \tau, \epsilon} \left[ \| \pi_\theta(\tau, a_t^\tau, \mathcal{C}) - (a_t - \epsilon) \|^2 \right],
                \label{eq:cfm_loss}
            \end{equation}
            where $\mathcal{C} = \{L, o^{vis}_t, s^{prop}_t, z\}$ denotes the conditioning context, comprising language instructions, visual observations, proprioception, and the inferred operator intent $z$.

%% file: Figure/framework.tex
\begin{figure*}[t!]
    \centering
    \includegraphics[width=1.0\linewidth]{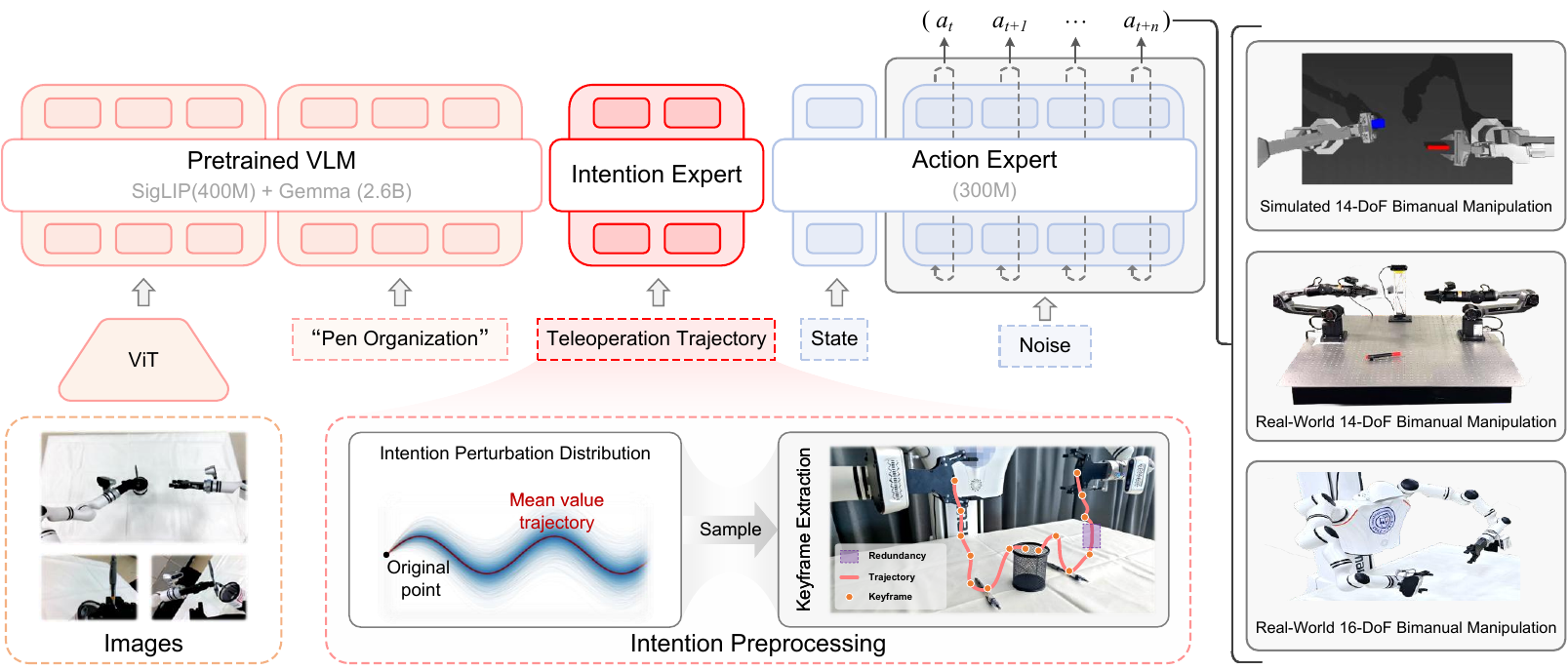}\\
    \caption{\textbf{Overview of the Adaptor framework.} The architecture comprises two primary phases: (i) Preprocessing, where    perturbation distributions and keyframes are extracted to model intent uncertainty; and (ii) Policy Learning. In this phase, the VLM backbone extracts environmental context, while the Intention Expert synthesizes semantic data with preprocessed trajectory guidance to infer latent intent, and the Action Expert employs flow matching to generate precise control commands.}
    \label{fig:framework}
\end{figure*}

%% file: Section/section5_experiment.tex
\section{Experiments} 

    We investigate two research questions: \textbf{RQ1}: Does Adaptor enhance manipulation efficiency and quality relative to pure teleoperation, and does it achieve state-of-the-art performance compared to assisted baselines? 
    -- Evaluated in Sec.~\ref{sec:comparative_evaluations} via multi-operator, multi-task comparisons utilizing both objective and subjective metrics. \textbf{RQ2}: What are the distinct contributions of each system component to the overall performance? -- Quantified in Sec.~\ref{sec:ablation_studies} via ablation studies isolating each component's effect.

    \subsection{Experimental Setup}

        \subsubsection{System Configurations and Benchmark}
    
            We evaluate our method on three robot platforms across six distinct manipulation tasks, as illustrated in Fig.~\ref{fig:task}.
            
            \noindent\textbf{ALOHA Simulation.} Based on the ALOHA setup~\cite{Zhao2023LearningFB}, this simulated environment features two 6-DoF Trossen ViperX arms and a single base camera (14-dimensional action space). We assess two bimanual tasks: \textit{Insertion}, which involves aligning a peg into a block, and \textit{Cube Transfer}, passing a cube from the right arm to the left.
            
            \noindent\textbf{AgileX PIPER.} This physical setup comprises two 6-DoF arms monitored by a four-camera system (two wrist, one base, and one low-angle). It operates within a 14-dimensional action space. The tasks include \textit{Pen Uncapping}, coordinating arms to remove a cap, and \textit{Shirt Folding}, folding a T-shirt initially laid flat.

            \noindent\textbf{Bimanual Realman.} A dual-arm platform equipped with two 7-DoF arms and three cameras (two wrist, one base), utilizing a 16-dimensional action space. We evaluate \textit{Pen Organization}, placing scattered pens into a holder, and \textit{Cube Stacking}, sequentially grasping and stacking cubes vertically.

            \noindent \textbf{Teleoperation Interface.} As illustrated in Fig.~\ref{fig:task} (right), the teleoperation system comprises a Head-Mounted Display (HMD) and handheld controllers. User inputs are mapped to the robot's joint space via Inverse Kinematics (IK). Simultaneously, concurrent multi-view video feeds are streamed to the HMD, facilitating immersive closed-loop control.

\input{Figure/task}

        \subsubsection{Baselines}
            We compare Adaptor against two baselines:
            
            \noindent\textbf{Full Teleop}~\cite{Zakka_Mink_Python_inverse_2025}. A direct control method where handheld end-effector poses are mapped to robot joint commands via inverse kinematics, relying solely on manual operator control without autonomous assistance.
            
            \noindent\textbf{HAJL}~\cite{luo2024human}. A shared control framework based on human-agent joint learning, employing a diffusion model to refine human inputs via reverse denoising.
            
        \subsubsection{Evaluation Metrics}
            We assess task performance using three metrics:
            
            \noindent\textbf{Teleoperation Quality.} The task success rate (\%), defined as the proportion of trials successfully completed within a fixed time limit.
            
            \noindent\textbf{Teleoperation Efficiency.} The completion time (s), defined as the duration of active human input (excluding autonomous execution time).
            
            \noindent\textbf{User Satisfaction.} Subjective feedback assessed using a questionnaire adapted from~\cite{liu2025casper}.

        \subsubsection{Participants and Procedures}
            Eleven healthy participants (3 females, 8 males) were recruited and provided written informed consent. Participants completed a practice session prior to performing the experimental tasks in a randomized order. A total of 30 trials were collected for each task-method combination. To simulate varying levels of operational proficiency, participants were randomly assigned to practice durations of 30, 60, or 120 minutes. Upon completion of each task, participants completed a user satisfaction questionnaire.

    \subsection{Comparative Evaluations}
    \label{sec:comparative_evaluations}
        \subsubsection{Comparative Results} 

        Table~\ref{tab:comparative_evaluations} presents a quantitative comparison of task success rates and teleoperation times between Adaptor and the baseline methods. Across all six evaluated tasks, Adaptor consistently achieves the highest success rate and the shortest teleoperation time. While model-assisted approaches (both HAJL~\cite{luo2024human} and Adaptor) generally outperform purely manual teleoperation (Full Teleop) in efficiency and efficacy, Adaptor demonstrates superior robustness. This is notably distinct from HAJL, which lacks explicit modeling of operator intent uncertainty and inter-operator variability, thereby limiting its generalization. Note that the reported teleoperation time strictly measures active human input, excluding the robot's autonomous execution. Unlike synchronous methods (Full Teleop, HAJL), Adaptor decouples demonstration from execution. This asynchronous design holds the potential to facilitate task queuing and Single-Operator-Multi-Robot (SOMR) workflows, allowing operators to efficiently manage parallel objectives.
        \input{Table/Table1}

        We attribute these performance gains to three strategic design choices. First, the intent-preprocessing module enhances recognition robustness via stochastic perturbations and keyframe extraction. Second, our random trajectory truncation strategy during training empowers the policy to operate effectively given only partial intent. This capability allows operators to provide concise demonstrations—often executing only a subset of the full trajectory—thereby significantly reducing teleoperation time. Finally, the integration of a Vision-Language-Action (VLA) backbone improves generalization, ensuring that the inferred intent accurately guides and reinforces end-to-end policy execution.

        \subsubsection{Cross-operator Generalization} 
            Table~\ref{tab:proficiency_analysis} presents the evaluation of system robustness across varying levels of operator proficiency (30, 60, and 120 minutes of practice). A positive correlation is observed between practice duration and success rate across all methods. This trend suggests that as operators gain experience, their control inputs increasingly converge towards the expert trajectory distribution, thereby facilitating task completion.

            However, Full Teleop and HAJL show performance variations correlated with operator proficiency. In the 30-minute trials, their success rates decrease to 38.43\% and 48.76\%, respectively, with higher standard deviations (22.08 and 19.20) reflecting increased variability. This decline is attributed to novice trajectories serving as out-of-distribution inputs, which hinders the baselines from mapping actions to the expert policy. In contrast, Adaptor (Ours) maintains a success rate of 83.21\% with a standard deviation of 5.69 under the same conditions. These results indicate that the proposed method minimizes the performance disparity between novice and expert users, improving cross-operator generalization.
            \input{Table/table2}

        \subsubsection{User Satisfaction Results} 
\begin{figure}[t]
                \centering
                \includegraphics[width=0.95\linewidth]{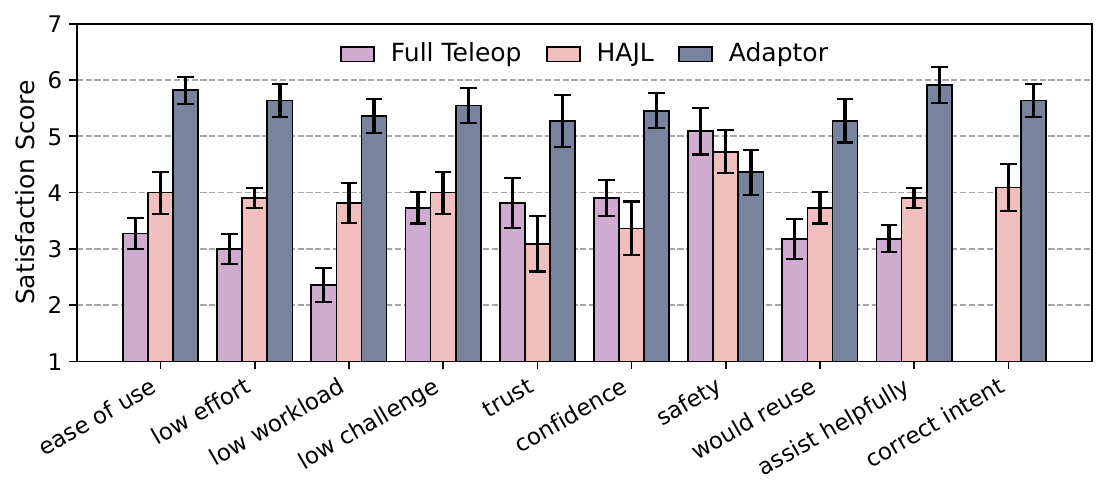}
                \caption{\textbf{Quantitative Analysis of User Satisfaction.} Mean satisfaction scores derived from questionnaires administered after participants completed 30 trials for each task--method combination. Data are averaged across all tasks, with error bars representing the standard deviation (SD).}
                \label{fig:Satisfaction}
            \end{figure}

            As illustrated in Fig.~\ref{fig:Satisfaction}, Adaptor demonstrates superior performance across nine of the ten metrics, with the notable exception of ``Perceived Safety''. This distinction stems from the fundamental difference in interaction paradigms. Adaptor operates via a one-shot demonstration followed by an autonomous execution workflow. While this paradigm substantially reduces operator workload, the absence of real-time intervention capabilities during the autonomous phase diminishes the operator's sense of agency. Consequently, this lowers perceived safety, reflecting a trade-off often observed in high-autonomy systems~\cite{collier2025sense}.

            In contrast, HAJL employs an iterative correction mechanism. However, its effectiveness is compromised by unstable intent inference and the need for frequent interruptions. These factors introduce temporal inefficiencies and degrade user experience, resulting in significantly lower scores for ``Trust'' and ``Willingness to Reuse''. Finally, while Full Teleop imposes the highest physical load due to the requirement for continuous manual input, its inherent real-time controllability fosters a higher sense of safety compared to semi-autonomous approaches.

    \subsection{Ablation Studies}
    \label{sec:ablation_studies}
        \subsubsection{Ablation Study on Noise Injection}
        To evaluate whether injecting noise into the intent representation improves trajectory diversity and generalization across operators, we conduct an ablation study on two tasks in the ALOHA simulation environment, as shown in Fig.~\ref{fig:ab_noise}~(left). The x-axis denotes the noise level (0 = no noise, i.e., the demonstrator trajectory is used directly as the intent), and the y-axis reports the task success rate. The results exhibit a clear inverted-U trend. With near-zero noise, the model overfits to the demonstrator's style and fails to generalize, yielding the lowest success rates. Conversely, excessive noise obscures the underlying intent and hinders recovery, also degrading performance. At a moderate noise level, the perturbation acts as an effective regularizer, encouraging task-relevant, operator-invariant intent representations and achieving the highest success.

        \subsubsection{Ablation Study on Keyframe Extraction} 
        
\begin{figure}[t]
            \centering
            \includegraphics[width=0.49\linewidth]{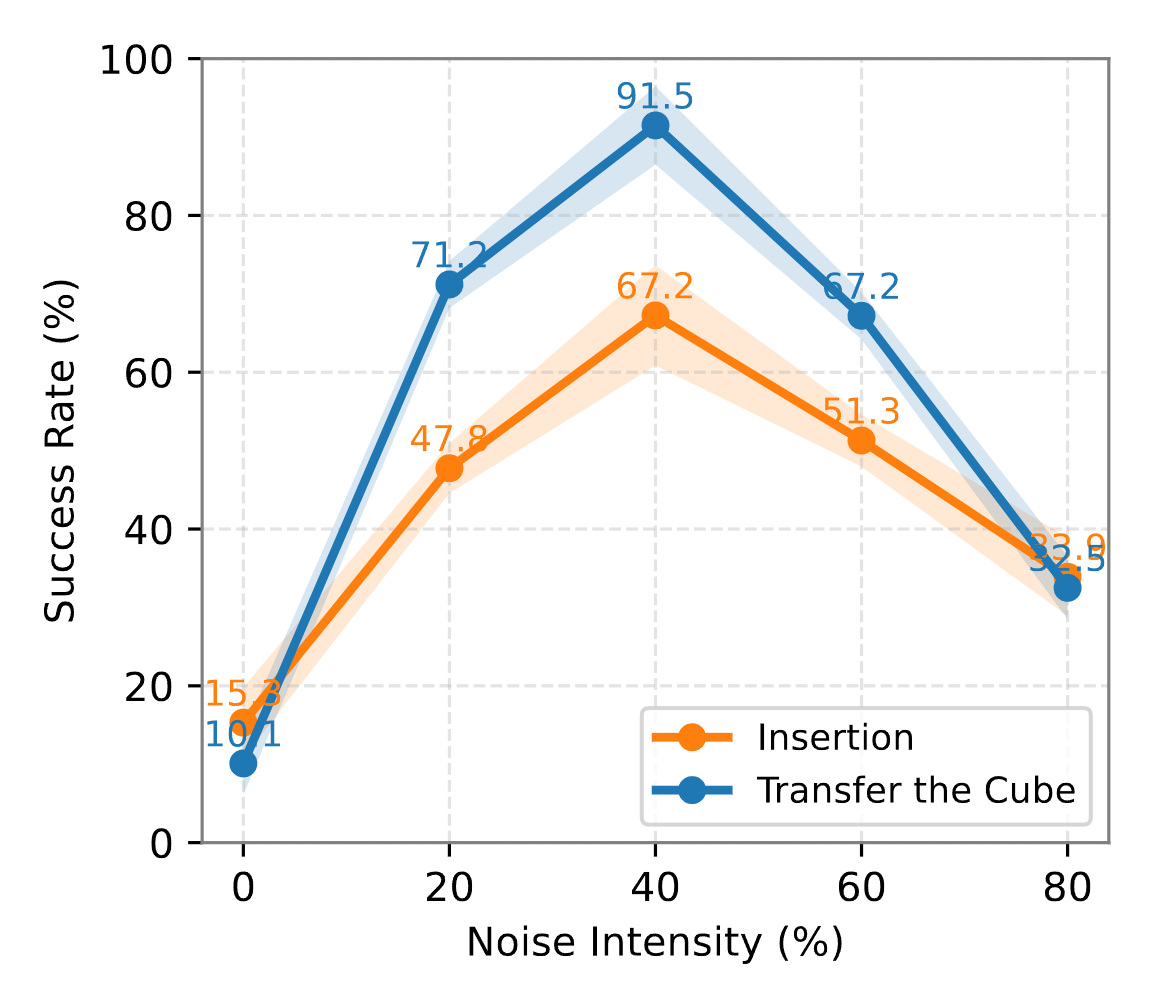}
            \hfill
            \includegraphics[width=0.49\linewidth]{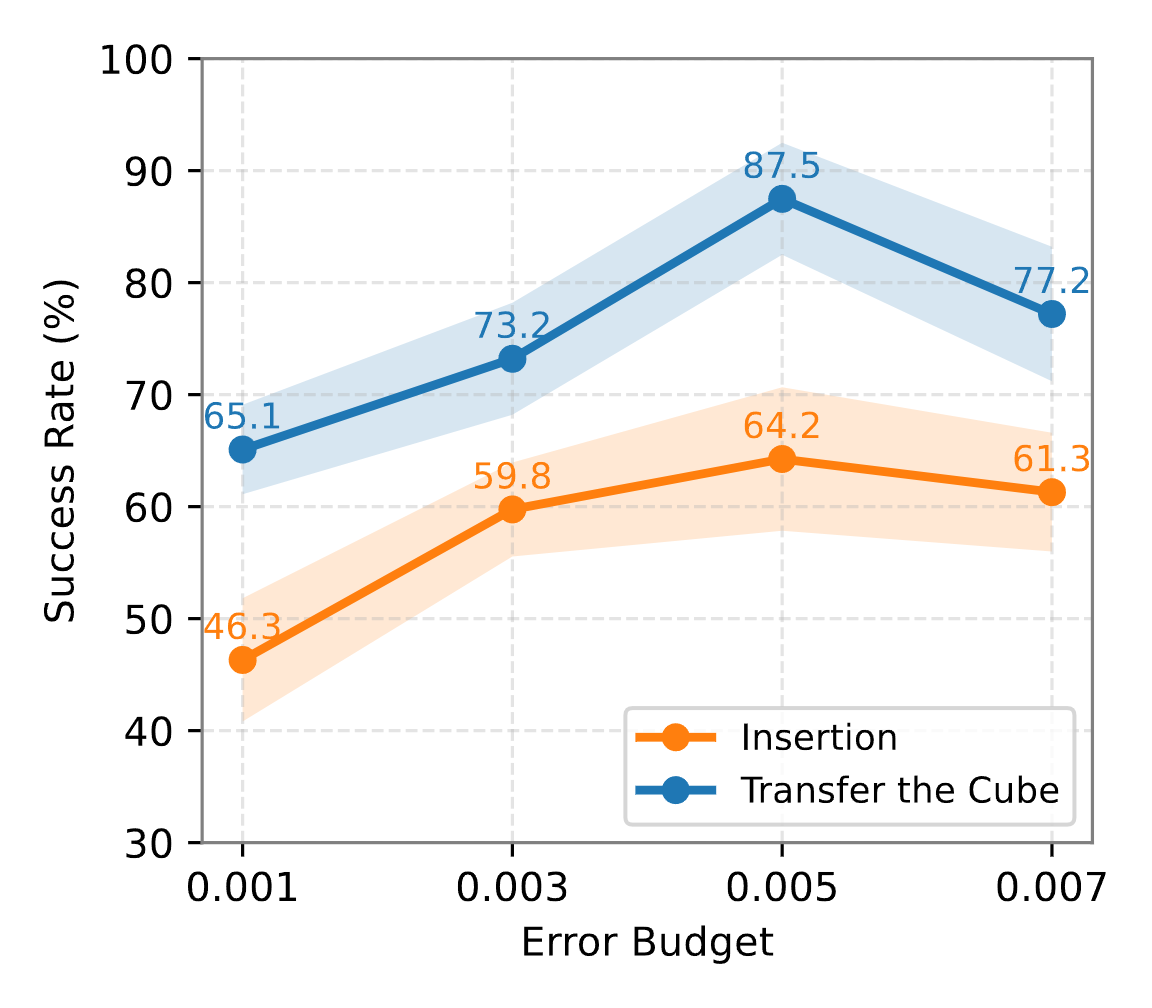}
            \caption{\textbf{Ablation study results}. Analysis of the proposed intent perturbation distribution and keyframe extraction on the Insertion and Cube Transfer tasks in the ALOHA simulation environment.}
            \label{fig:ab_noise}
        \end{figure}

            Fig.~\ref{fig:ab_noise}~(right) presents the success rates of two tasks in the ALOHA simulation environment as a function of trajectory reconstruction error. The horizontal axis denotes the reconstruction error (smaller values indicate a closer match between the reconstructed and original trajectories, i.e., denser keyframes), and the vertical axis denotes the task success rate. Both tasks exhibit a characteristic inverted-U relationship: when the Error Budget ($\eta$) is too large, the extracted keyframes are overly sparse, yielding an underspecified trajectory representation that fails to capture the user's latent intent; conversely, when $\eta$ is too small, excessively dense keyframes introduce redundant waypoints and accumulate noise, increasing planning and control overhead and inducing overfitting, which likewise reduces success rates. Overall, a moderate $\eta$ strikes a balance between representational fidelity and generalization, producing the highest success rates on both tasks.

%% file: Figure/task.tex
\begin{figure*}
    \centering
    \includegraphics[width=1.0\linewidth]{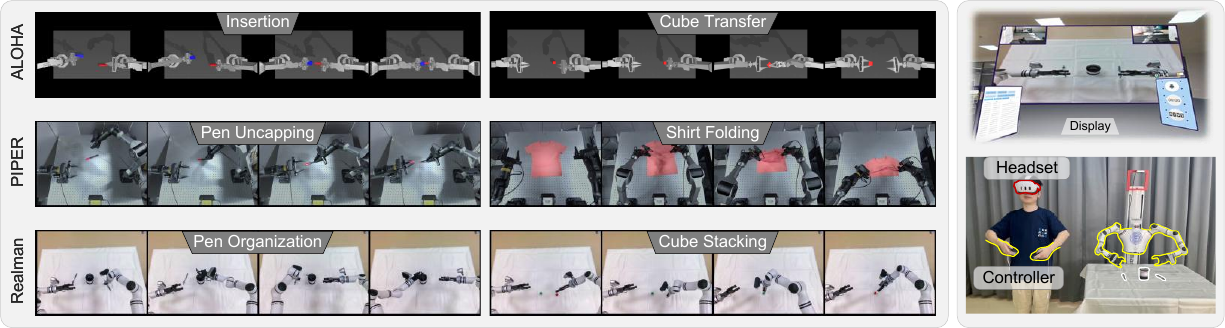}
    \caption{\textbf{Overview of the experimental setup.} Left: Representative tasks across three robotic platforms, including ALOHA simulator (Insertion, Cube Transfer), PIPER (Pen Uncapping, Shirt Folding), and Realman (Pen Organization, Cube Stacking). Right: Schematic of the teleoperation system architecture.}
    \label{fig:task}
\end{figure*}

%% file: Table/Table1.tex
\begin{table*}[t!]
  \centering
  \begin{threeparttable}
    \caption{Quantitative Evaluation on Real-World and Simulated Benchmarks.}
    \label{tab:comparative_evaluations}
    
    \setlength{\tabcolsep}{7.5pt} 
    \renewcommand{\arraystretch}{1.3} 
    
    \begin{tabular}{ll cc >{\columncolor{lightredbg}}c c cc >{\columncolor{lightredbg}}c c}
      \toprule
      \multirow{2.5}{*}{\textbf{Robot}} & \multirow{2.5}{*}{\textbf{Task}} &
      \multicolumn{4}{c}{\textbf{Success Rate} (\%, $\uparrow$)} &
      \multicolumn{4}{c}{\textbf{Teleoperation Time} (s, $\downarrow$)} \\
      \cmidrule(r){3-6} \cmidrule(l){7-10}
       & & Full Teleop~\cite{Zakka_Mink_Python_inverse_2025} & HAJL~\cite{luo2024human} & \textbf{Adaptor} & Improv. & Full Teleop~\cite{Zakka_Mink_Python_inverse_2025} & HAJL~\cite{luo2024human} & \textbf{Adaptor} & Improv. \\
      \midrule
      
      \multirow{2}{*}{ALOHA} 
        & Insertion & 33.27 & 41.83 & \textbf{67.25} & \textcolor{green!60!black}{+60.77\%} 
        & 38.77 & 32.29 & \textbf{17.92} & \textcolor{green!60!black}{-44.50\%} 
        \\
        & Cube Transfer & 63.64 & 84.64 & \textbf{91.48} & \textcolor{green!60!black}{+8.08\%} 
        & 24.23 & 19.38 & \textbf{11.87} & \textcolor{green!60!black}{-38.75\%} 
        \\
      \cmidrule{1-10} 
      
      \multirow{2}{*}{PIPER}  
        & Pen Uncapping & 45.09 & 60.75 & \textbf{89.36} & \textcolor{green!60!black}{+47.09\%} 
        & 20.36 & 15.59 & \textbf{10.39} & \textcolor{green!60!black}{-33.35\%} 
        \\
        & Shirt Folding & 38.00 & 47.24 & \textbf{76.10} & \textcolor{green!60!black}{+61.09\%} 
        & 23.54 & 15.73 & \textbf{11.59} & \textcolor{green!60!black}{-26.32\%} 
        \\
      \cmidrule{1-10}
      
      \multirow{2}{*}{Realman} 
        & Pen Org. & 42.91 & 59.92 & \textbf{90.27} & \textcolor{green!60!black}{+50.65\%} 
        & 31.15 & 24.49 & \textbf{17.50} & \textcolor{green!60!black}{-28.54\%} 
        \\
        & Cube Stacking & 55.63 & 69.51 & \textbf{86.07} & \textcolor{green!60!black}{+23.82\%} 
        & 16.13 & 13.18 & \textbf{10.34} & \textcolor{green!60!black}{-21.55\%} \\
      \bottomrule
    \end{tabular}

    \begin{tablenotes}[para, flushleft] 
      \footnotesize
      \setlength{\fboxsep}{1.5pt} 
       \emph{Notes:} We report the average Success Rate ($\uparrow$) and Teleoperation Time ($\downarrow$). Our method (Adaptor) is highlighted in \colorbox{lightredbg}{red}; ``Improv.'' denotes  the  relative improvement (\%) over the SOTA baseline. The best results are in \textbf{bold}.
    \end{tablenotes}
    
  \end{threeparttable}
  \vspace{-3mm} 
\end{table*}

%% file: Table/table2.tex
\begin{table}[t!]
  \centering
  \small
  \begin{threeparttable}
    \caption{Cross-operator Generalization Analysis.}
    \label{tab:proficiency_analysis}
    
    \small 
    \setlength{\tabcolsep}{7pt} 
    \renewcommand{\arraystretch}{1.2}

    \begin{tabular}{l c c c c}
        \toprule
        \multirow{2}{*}{\textbf{Method}} & \multicolumn{3}{c}{\textbf{Success Rate} (\%) $\uparrow$} &\multirow{2}{*}{\textbf{SD} $\downarrow$}  \\
        \cmidrule(lr){2-4} 
         & \textbf{30 min} & \textbf{60 min} & \textbf{120 min} & \\
        \midrule
        
        Full Teleop~\cite{Zakka_Mink_Python_inverse_2025} 
        & 38.43 & 62.17 & 82.54 & 22.08 \\
        
        HAJL~\cite{luo2024human} 
        & 48.76 & 71.32 & 86.95 & 19.20 \\
        
        \addlinespace[3pt]
        \rowcolor{lightredbg}
        \textbf{Adaptor (Ours)} 
        & \textbf{83.21} & \textbf{89.85} & \textbf{94.53} & \textbf{5.69} \\
        \bottomrule
    \end{tabular}

    \begin{tablenotes}[para, flushleft]
      \footnotesize
      \setlength{\fboxsep}{1.5pt} 
      \emph{Notes:}
      Success Rate (\%) is averaged across three proficiency levels for each practice duration. The SD serves to quantify generalization, where a lower value indicates superior consistency across these levels.
    \end{tablenotes}
  \end{threeparttable}
  
\vspace{-3mm} 
    
\end{table}

%% file: Section/section6_conclusion.tex
\section{Conclusion}

    We present Adaptor, an assistive teleoperation framework designed to achieve intent generalization across diverse operators. By integrating intent uncertainty simulation during the preprocessing phase with explicit intent modeling in end-to-end training, Adaptor effectively addresses the challenge of recognizing out-of-distribution (OOD) intents stemming from long-tail operator behaviors. Extensive evaluations across six manipulation tasks demonstrate that Adaptor significantly outperforms state-of-the-art (SOTA) methods, achieving a 41.92\% increase in average success rate and a 32.17\% reduction in teleoperation time. Furthermore, a significant reduction in the standard deviation of success rates across operators of varying proficiency validates the system's superior cross-operator generalization capabilities. User studies further confirm that our approach yields substantially higher user satisfaction.

    While Adaptor outperforms assisted teleoperation baselines across varying experience levels, intent recognition remains susceptible to errors when operator behavior diverges significantly from the demonstrator's distribution (e.g., with untrained novices). Furthermore, although our assistance operates autonomously to markedly reduce operator workload, post-study feedback indicates that concerns regarding system safety persist. In future work, we propose adopting a hybrid supervised-reinforcement fine-tuning strategy designed to facilitate controlled exploration of out-of-distribution (OOD) intents. This will allow us to strictly bolster safety and trust while maintaining minimal operator intervention.

%% file: ref.bib
@article{10035484,
  author  = {Darvish, Kourosh and Penco, Luigi and Ramos, Joao and Cisneros, Rafael and Pratt, Jerry and Yoshida, Eiichi and Ivaldi, Serena and Pucci, Daniele},
  title   = {Teleoperation of Humanoid Robots: A Survey},
  journal = {IEEE Transactions on Robotics},
  year    = {2023},
  volume  = {39},
  number  = {3},
  pages   = {1706--1727}
}

@article{luo2024human,
  title={Human-agent joint learning for efficient robot manipulation skill acquisition},
  author={Luo, Shengcheng and Peng, Quanquan and Lv, Jun and Hong, Kaiwen and Driggs-Campbell, Katherine Rose and Lu, Cewu and Li, Yong-Lu},
  journal={arXiv preprint arXiv:2407.00299},
  year={2024}
}

@article{black2024pi_0,
  title={{$\pi_0$}: A Vision-Language-Action Flow Model for General Robot Control},
  author={Black, Kevin and Brown, Noah and Driess, Danny and Esmail, Adnan and Equi, Michael and Finn, Chelsea and Fusai, Niccolo and Groom, Lachy and Hausman, Karol and Ichter, Brian and others},
  journal={arXiv preprint arXiv:2410.24164},
  year={2024}
}

@article{liu2024rdt,
  title={Rdt-1b: a diffusion foundation model for bimanual manipulation},
  author={Liu, Songming and Wu, Lingxuan and Li, Bangguo and Tan, Hengkai and Chen, Huayu and Wang, Zhengyi and Xu, Ke and Su, Hang and Zhu, Jun},
  journal={arXiv preprint arXiv:2410.07864},
  year={2024}
}

@article{kim2024openvla,
  title={Openvla: An open-source vision-language-action model},
  author={Kim, Moo Jin and Pertsch, Karl and Karamcheti, Siddharth and Xiao, Ted and Balakrishna, Ashwin and Nair, Suraj and Rafailov, Rafael and Foster, Ethan and Lam, Grace and Sanketi, Pannag and others},
  journal={arXiv preprint arXiv:2406.09246},
  year={2024}
}

@article{iyer2024open,
  title={Open teach: A versatile teleoperation system for robotic manipulation},
  author={Iyer, Aadhithya and Peng, Zhuoran and Dai, Yinlong and Guzey, Irmak and Haldar, Siddhant and Chintala, Soumith and Pinto, Lerrel},
  journal={arXiv preprint arXiv:2403.07870},
  year={2024}
}

@misc{Zakka_Mink_Python_inverse_2025,
  author = {Zakka, Kevin},
  title = {{Mink: Python inverse kinematics based on MuJoCo}},
  year = {2026},
  month = feb,
  version = {1.1.0},
  url = {https://github.com/kevinzakka/mink},
  license = {Apache-2.0}
}

@article{10.1109/TRO.2024.3431830,
  author  = {Phung, Amy and Billings, Gideon and Daniele, Andrea F. and Walter, Matthew R. and Camilli, Richard},
  title   = {A Shared Autonomy System for Precise and Efficient Remote Underwater Manipulation},
  journal = {IEEE Transactions on Robotics},
  year    = {2024},
  volume  = {40},
  month   = jan,
  pages   = {4147--4159},
  doi     = {10.1109/TRO.2024.3431830}
}

@article{yoneda2023noise,
  title={To the noise and back: Diffusion for shared autonomy},
  author={Yoneda, Takuma and Sun, Luzhe and Yang, Ge and Stadie, Bradly and Walter, Matthew},
  journal={arXiv preprint arXiv:2302.12244},
  year={2023}
}

@inproceedings{padmanabha2024independence,
  title={Independence in the home: A wearable interface for a person with quadriplegia to teleoperate a mobile manipulator},
  author={Padmanabha, Akhil and Gupta, Janavi and Chen, Chen and Yang, Jehan and Nguyen, Vy and Weber, Douglas J and Majidi, Carmel and Erickson, Zackory},
  booktitle={Proceedings of the 2024 ACM/IEEE International Conference on Human-Robot Interaction},
  pages={542--551},
  year={2024}
}

@article{liu2024dragon,
  title={Dragon: A dialogue-based robot for assistive navigation with visual language grounding},
  author={Liu, Shuijing and Hasan, Aamir and Hong, Kaiwen and Wang, Runxuan and Chang, Peixin and Mizrachi, Zachary and Lin, Justin and McPherson, D Livingston and Rogers, Wendy A and Driggs-Campbell, Katherine},
  journal={IEEE Robotics and Automation Letters},
  volume={9},
  number={4},
  pages={3712--3719},
  year={2024},
  publisher={IEEE}
}

@inproceedings{chen2022asha,
  title={Asha: Assistive teleoperation via human-in-the-loop reinforcement learning},
  author={Chen, Sean and Gao, Jensen and Reddy, Siddharth and Berseth, Glen and Dragan, Anca D and Levine, Sergey},
  booktitle={2022 International Conference on Robotics and Automation (ICRA)},
  pages={7505--7512},
  year={2022},
  organization={IEEE}
}

@article{selvaggio2021autonomy,
  title={Autonomy in physical human-robot interaction: A brief survey},
  author={Selvaggio, Mario and Cognetti, Marco and Nikolaidis, Stefanos and Ivaldi, Serena and Siciliano, Bruno},
  journal={IEEE Robotics and Automation Letters},
  volume={6},
  number={4},
  pages={7989--7996},
  year={2021},
  publisher={IEEE}
}

@article{gu2023rt,
  title={Rt-trajectory: Robotic task generalization via hindsight trajectory sketches},
  author={Gu, Jiayuan and Kirmani, Sean and Wohlhart, Paul and Lu, Yao and Arenas, Montserrat Gonzalez and Rao, Kanishka and Yu, Wenhao and Fu, Chuyuan and Gopalakrishnan, Keerthana and Xu, Zhuo and others},
  journal={arXiv preprint arXiv:2311.01977},
  year={2023}
}

@inproceedings{sundaresan2023rt,
  title={Rt-sketch: Goal-conditioned imitation learning from hand-drawn sketches},
  author={Sundaresan, Priya and Vuong, Quan and Gu, Jiayuan and Xu, Peng and Xiao, Ted and Kirmani, Sean and Yu, Tianhe and Stark, Michael and Jain, Ajinkya and Hausman, Karol and others},
  booktitle={8th Annual Conference on Robot Learning},
  year={2024}
}

@article{hoffman2024inferring,
  title={Inferring human intent and predicting human action in human--robot collaboration},
  author={Hoffman, Guy and Bhattacharjee, Tapomayukh and Nikolaidis, Stefanos},
  journal={Annual Review of Control, Robotics, and Autonomous Systems},
  volume={7},
  number={1},
  pages={73--95},
  year={2024},
  publisher={Annual Reviews}
}

@article{liu2025casper,
  title={Casper: Inferring Diverse Intents for Assistive Teleoperation with Vision Language Models},
  author={Liu, Huihan and Shah, Rutav and Liu, Shuijing and Pittenger, Jack and Seo, Mingyo and Cui, Yuchen and Bisk, Yonatan and Mart{\'\i}n-Mart{\'\i}n, Roberto and Zhu, Yuke},
  journal={arXiv preprint arXiv:2506.14727},
  year={2025}
}

@INPROCEEDINGS{brooks2019,
  author={Brooks, Connor and Szafir, Daniel},
  booktitle={2019 14th ACM/IEEE International Conference on Human-Robot Interaction (HRI)}, 
  title={Balanced Information Gathering and Goal-Oriented Actions in Shared Autonomy}, 
  year={2019},
  pages={85--94}}

@article{10.1145/3359614,
  author    = {Jain, Siddarth and Argall, Brenna},
  title     = {Probabilistic Human Intent Recognition for Shared Autonomy in Assistive Robotics},
  journal   = {ACM Transactions on Human-Robot Interaction},
  year      = {2020},
  volume    = {9},
  number    = {1},
  articleno = {2},
  numpages  = {23},
  doi       = {10.1145/3359614}
}

@article{javdani2018shared,
  title={Shared autonomy via hindsight optimization for teleoperation and teaming},
  author={Javdani, Shervin and Admoni, Henny and Pellegrinelli, Stefania and Srinivasa, Siddhartha S and Bagnell, J Andrew},
  journal={The International Journal of Robotics Research},
  volume={37},
  number={7},
  pages={717--742},
  year={2018},
  publisher={SAGE Publications Sage UK: London, England}
}

@inproceedings{jonnavittula2021know,
  title={I know what you meant: Learning human objectives by (under) estimating their choice set},
  author={Jonnavittula, Ananth and Losey, Dylan P},
  booktitle={2021 IEEE International Conference on Robotics and Automation (ICRA)},
  pages={2747--2753},
  year={2021},
  organization={IEEE}
}

@article{newman2022harmonic,
  title={{HARMONIC}: A multimodal dataset of assistive human--robot collaboration},
  author={Newman, Benjamin A and Aronson, Reuben M and Srinivasa, Siddhartha S and Kitani, Kris and Admoni, Henny},
  journal={The International Journal of Robotics Research},
  volume={41},
  number={1},
  pages={3--11},
  year={2022},
  publisher={SAGE Publications Sage UK: London, England}
}

@article{zhao2024conformalized,
  title={Conformalized teleoperation: Confidently mapping human inputs to high-dimensional robot actions},
  author={Zhao, Michelle and Simmons, Reid and Admoni, Henny and Bajcsy, Andrea},
  journal={arXiv preprint arXiv:2406.07767},
  year={2024}
}

@inproceedings{cui2023no,
  title={No, to the right: Online language corrections for robotic manipulation via shared autonomy},
  author={Cui, Yuchen and Karamcheti, Siddharth and Palleti, Raj and Shivakumar, Nidhya and Liang, Percy and Sadigh, Dorsa},
  booktitle={Proceedings of the 2023 ACM/IEEE International Conference on Human-Robot Interaction},
  pages={93--101},
  year={2023}
}

@inproceedings{Jonnavittula2022LearningTS,
  author={Jonnavittula, Ananth and Losey, Dylan P.},
  booktitle={2021 IEEE/RSJ International Conference on Intelligent Robots and Systems (IROS)}, 
  title={Learning to Share Autonomy Across Repeated Interaction}, 
  year={2021},
  volume={},
  number={},
  pages={1851-1858},
}

@inproceedings{zurek2021situational,
  title={Situational confidence assistance for lifelong shared autonomy},
  author={Zurek, Matthew and Bobu, Andreea and Brown, Daniel S and Dragan, Anca D},
  booktitle={2021 IEEE International Conference on Robotics and Automation (ICRA)},
  pages={2783--2789},
  year={2021},
  organization={IEEE}
}

@article{jonnavittula2024sari,
  title={SARI: Shared autonomy across repeated interaction},
  author={Jonnavittula, Ananth and Mehta, Shaunak A and Losey, Dylan P},
  journal={ACM Transactions on Human-Robot Interaction},
  volume={13},
  number={2},
  pages={1--36},
  year={2024},
  publisher={ACM New York, NY}
}

@article{mehta2025stable,
  title={Stable-BC: Controlling covariate shift with stable behavior cloning},
  author={Mehta, Shaunak A and Ciftci, Yusuf Umut and Ramachandran, Balamurugan and Bansal, Somil and Losey, Dylan P},
  journal={IEEE Robotics and Automation Letters},
  year={2025},
  publisher={IEEE}
}

@inproceedings{baek2024unexplored,
  title={Unexplored faces of robustness and out-of-distribution: Covariate shifts in environment and sensor domains},
  author={Baek, Eunsu and Park, Keondo and Kim, Jiyoon and Kim, Hyung-Sin},
  booktitle={Proceedings of the IEEE/CVF Conference on Computer Vision and Pattern Recognition},
  pages={22294--22303},
  year={2024}
}

@inproceedings{laskey2017dart,
  title={Dart: Noise injection for robust imitation learning},
  author={Laskey, Michael and Lee, Jonathan and Fox, Roy and Dragan, Anca and Goldberg, Ken},
  booktitle={Conference on robot learning},
  pages={143--156},
  year={2017},
  organization={PMLR}
}

@article{liu2024robust,
  title={Robust adaptive control of high-order fully-actuated systems: Command filtered backstepping with concurrent learning},
  author={Liu, Weizhen and Duan, Guangren and Hou, Mingzhe and Kong, He},
  journal={IEEE Transactions on Circuits and Systems I: Regular Papers},
  volume={71},
  number={12},
  pages={5780--5791},
  year={2024},
  publisher={IEEE}
}

@article{GREEN1986351,
  title={Persistence of excitation in linear systems},
  author={Green, Michael and Moore, John B},
  journal={Systems \& control letters},
  volume={7},
  number={5},
  pages={351--360},
  year={1986},
  publisher={Elsevier}
}

@inproceedings{kelly2019hg,
  title={Hg-dagger: Interactive imitation learning with human experts},
  author={Kelly, Michael and Sidrane, Chelsea and Driggs-Campbell, Katherine and Kochenderfer, Mykel J},
  booktitle={2019 International Conference on Robotics and Automation (ICRA)},
  pages={8077--8083},
  year={2019},
  organization={IEEE}
}

@inproceedings{wang2025inference,
  title={Inference-time policy steering through human interactions},
  author={Wang, Yanwei and Wang, Lirui and Du, Yilun and Sundaralingam, Balakumar and Yang, Xuning and Chao, Yu-Wei and P{\'e}rez-D’Arpino, Claudia and Fox, Dieter and Shah, Julie},
  booktitle={2025 IEEE International Conference on Robotics and Automation (ICRA)},
  pages={15626--15633},
  year={2025},
  organization={IEEE}
}

@inproceedings{radford2021learning,
  title={Learning transferable visual models from natural language supervision},
  author={Radford, Alec and Kim, Jong Wook and Hallacy, Chris and Ramesh, Aditya and Goh, Gabriel and Agarwal, Sandhini and Sastry, Girish and Askell, Amanda and Mishkin, Pamela and Clark, Jack and others},
  booktitle={International conference on machine learning},
  pages={8748--8763},
  year={2021},
  organization={PmLR}
}

@inproceedings{li2023blip,
  title={Blip-2: Bootstrapping language-image pre-training with frozen image encoders and large language models},
  author={Li, Junnan and Li, Dongxu and Savarese, Silvio and Hoi, Steven},
  booktitle={International conference on machine learning},
  pages={19730--19742},
  year={2023},
  organization={PMLR}
}

@article{gao2023llama,
  title={Llama-adapter v2: Parameter-efficient visual instruction model},
  author={Gao, Peng and Han, Jiaming and Zhang, Renrui and Lin, Ziyi and Geng, Shijie and Zhou, Aojun and Zhang, Wei and Lu, Pan and He, Conghui and Yue, Xiangyu and others},
  journal={arXiv preprint arXiv:2304.15010},
  year={2023}
}

@article{intelligence2025-pi05,
  title   = {$\pi_{0.5}$: A Vision-Language-Action Model with Open-World Generalization},
  author  = {{Physical Intelligence} and Black, Kevin and Brown, Noah and Darpinian, James and Dhabalia, Karan and Driess, Danny and Esmail, Adnan and Equi, Michael and Finn, Chelsea and Fusai, Niccolo and others},
  journal = {arXiv preprint arXiv:2504.16054},
  year    = {2025}
}

@inproceedings{zitkovich2023rt,
  title={Rt-2: Vision-language-action models transfer web knowledge to robotic control},
  author={Zitkovich, Brianna and Yu, Tianhe and Xu, Sichun and Xu, Peng and Xiao, Ted and Xia, Fei and Wu, Jialin and Wohlhart, Paul and Welker, Stefan and Wahid, Ayzaan and others},
  booktitle={Conference on Robot Learning},
  pages={2165--2183},
  year={2023},
  organization={PMLR}
}

@article{liu2025hybridvla,
  title={Hybridvla: Collaborative diffusion and autoregression in a unified vision-language-action model},
  author={Liu, Jiaming and Chen, Hao and An, Pengju and Liu, Zhuoyang and Zhang, Renrui and Gu, Chenyang and Li, Xiaoqi and Guo, Ziyu and Chen, Sixiang and Liu, Mengzhen and others},
  journal={arXiv preprint arXiv:2503.10631},
  year={2025}
}

@article{Zhao2023LearningFB,
    title={Learning Fine-Grained Bimanual Manipulation with Low-Cost Hardware},
    author={Tony Zhao and Vikash Kumar and Sergey Levine and Chelsea Finn},
    journal={RSS},
    year={2023},
    volume={abs/2304.13705},
}

@inproceedings{ross2010efficient,
  title={Efficient reductions for imitation learning},
  author={Ross, St{\'e}phane and Bagnell, Drew},
  booktitle={Proceedings of the thirteenth international conference on artificial intelligence and statistics},
  pages={661--668},
  year={2010},
  organization={JMLR Workshop and Conference Proceedings}
}

@inproceedings{ross2011reduction,
  title={A reduction of imitation learning and structured prediction to no-regret online learning},
  author={Ross, St{\'e}phane and Gordon, Geoffrey and Bagnell, Drew},
  booktitle={Proceedings of the fourteenth international conference on artificial intelligence and statistics},
  pages={627--635},
  year={2011},
  organization={JMLR Workshop and Conference Proceedings}
}

@article{hu2022lora,
  title={Lora: Low-rank adaptation of large language models.},
  author={Hu, Edward J and Shen, Yelong and Wallis, Phillip and Allen-Zhu, Zeyuan and Li, Yuanzhi and Wang, Shean and Wang, Liang and Chen, Weizhu and others},
  journal={Iclr},
  volume={1},
  number={2},
  pages={3},
  year={2022}
}

@misc{team2025gemma,
      title={Gemma 3 Technical Report}, 
      author={Gemma Team and Aishwarya Kamath and Johan Ferret and Shreya Pathak and Nino Vieillard and Ramona Merhej and Sarah Perrin and Tatiana Matejovicova and Alexandre Ramé and Morgane Rivière and others},
      year={2025},
      eprint={2503.19786},
      archivePrefix={arXiv},
      primaryClass={cs.CL},
      url={https://arxiv.org/abs/2503.19786}, 
}

@inproceedings{collier2025sense,
  title={The sense of agency in assistive robotics using shared autonomy},
  author={Collier, Maggie A and Narayan, Rithika and Admoni, Henny},
  booktitle={2025 20th ACM/IEEE International Conference on Human-Robot Interaction (HRI)},
  pages={880--888},
  year={2025},
  organization={IEEE}
}
